\documentclass[journal]{IEEEtran}
\usepackage{amsmath,amsfonts}
\usepackage{algorithmic}
\usepackage{algorithm}
\usepackage{array}
\usepackage[caption=false,font=normalsize,labelfont=sf,textfont=sf]{subfig}
\usepackage{textcomp}
\usepackage{stfloats}
\usepackage{url}
\usepackage{verbatim}
\usepackage{graphicx}
\usepackage{cite}
\usepackage[dvipsnames]{xcolor}
\usepackage{xcolor}
\usepackage{lmodern}
\usepackage{comment}

\usepackage{hyperref}
\hypersetup{
    colorlinks=true,
    citecolor=black,   
    linkcolor=blue,    
    urlcolor=blue      
}

\begin{document}

\title{Real-Time Buoyancy Estimation for AUV Simulations Using Convex Hull-Based Submerged Volume Calculation}


\author{\IEEEauthorblockA{Ad-Deen Mahbub\IEEEauthorrefmark{1},
 and Md Ragib Shaharear\IEEEauthorrefmark{2}\\}
\IEEEauthorblockA{\IEEEauthorrefmark{1} Department of Electrical \& Electronic Engineering, Shahjalal University of Science \& Technology \\
}
\IEEEauthorblockA{\IEEEauthorrefmark{2}Department of Electrical \& Computer Engineering, Johns Hopkins University
}}

\markboth{AQ²UASIM: Advancing Quantitative and QUAlitative SIMulators for marine applications, ICRA(2025)}%
{Shell \MakeLowercase{\textit{et al.}}: ViT NAS}

\IEEEpubid{}

\maketitle

\begin{abstract}
Accurate real-time buoyancy modeling is essential for high-fidelity Autonomous Underwater Vehicle (AUV) simulations, yet NVIDIA Isaac Sim lacks a native buoyancy system, requiring external solutions for precise underwater physics. This paper presents a novel convex hull-based approach to dynamically compute the submerged volume of an AUV in real time. By extracting mesh geometry from the simulation environment and calculating the hull portion intersecting the water level along the z-axis, our method enhances accuracy over traditional geometric approximations. A cross-sectional area extension reduces computational overhead, enabling efficient buoyant force updates that adapt to orientation, depth, and sinusoidal wave fluctuations (±0.3 m). Tested on a custom AUV design for SAUVC 2025, this approach delivers real-time performance and scalability, improving simulation fidelity for underwater robotics research without precomputed hydrodynamic models.
\end{abstract}




\section{Introduction}

Buoyancy, an upward force governed by Archimedes’ principle, is fundamental to the behavior of Autonomous Underwater Vehicles (AUVs) and surface vessels, influencing stability, maneuverability, and hydrodynamic performance. Traditional buoyancy models in robotic simulation platforms, such as Gazebo and NVIDIA Isaac Sim, lack built-in systems for body volume submersion calculations and instead rely on simplified geometric approximations, modeling vehicles as basic shapes like cylinders, spheres, or cuboids \cite{c1}. Although these methods offer computational simplicity, they fail to capture the irregular hull geometries of modern AUV designs and the dynamic water interactions that significantly affect submerged volume and, consequently, buoyant force \cite{c2}. For instance, computational fluid dynamics (CFD) studies reveal that hull shape complexity impacts hydrodynamic accuracy \cite{c2}, yet their computational cost renders them impractical for real-time applications \cite{c7}. This limitation compromises the fidelity of simulations used for AUV control system development and mission planning.
To address these shortcomings, we propose a real-time buoyancy computation framework that leverages convex hull algorithms \cite{c4} and mesh data relationships to precisely determine an AUV’s submerged volume at any instant. Our approach extracts the AUV’s mesh geometry from the simulation environment and iteratively computes the hull portion intersecting the water level along the z-axis, adapting dynamically to changes in depth, orientation, and external forces. Unlike static convex hull methods applied in other domains \cite{c3,c4}, our framework incorporates varying water levels, including surface wave effects, enhancing realism for surface vessels and semi-submerged AUV operations. This capability aligns with prior work on environmental influences, such as ice surface proximity \cite{c7}, but extends it to real-time contexts. By continuously updating submerged volume based on water interactions, our method delivers high-fidelity simulations suitable for control design, mission planning, and performance evaluation of underwater robots.

This paper details the implementation of our real-time buoyancy computation model and assesses its accuracy against conventional approaches \cite{c1}. Our key contributions are:

\begin{itemize}
    \item Proposed an approach that dynamically computes the submerged volume of an AUV using convex hull algorithms and mesh-based calculations.
    \item Incorporating water surface interactions and wave effects to improve the realism of underwater and surface vehicle simulations.
    \item Development of an optimised algorithm suitable for real-time execution in NVIDIA Isaac Sim and other robotics simulation platforms.

\end{itemize}

\section{Related Work}

Robotics simulation and hardware implementation of autonomous robotics \cite{c16,c17,c18,c19} are major research areas in the field of automation \cite{c20}. Buoyancy simulation is vital for modeling Autonomous Underwater Vehicles (AUVs) and surface vessels in platforms like Gazebo and NVIDIA Isaac Sim. Traditional approaches approximate AUVs as simple geometric shapes (e.g., cylinders or ellipsoids) to estimate submerged volume and buoyant force, but they falter with irregular hulls and dynamic water interactions \cite{c1}. Computational fluid dynamics (CFD) has been used to optimize AUV hull shapes, providing detailed volume and drag estimates \cite{c2}, though its computational intensity hinders real-time use. Convex hull algorithms offer geometric precision for volume computation, such as incremental hull processing for 3D point clouds \cite{c3} and the Quickhull algorithm for static mesh objects \cite{c4}, yet these lack fluid dynamics integration. Buoyancy control systems adjust volume via electromechanical modules \cite{c5} or identify residual buoyancy in real-time \cite{c6}, but they rely on predefined geometries rather than dynamic mesh updates.

Further work has explored mesh-based hydrodynamics and real-time simulation. CFD studies near environmental boundaries (e.g., ice surfaces) reveal volume impacts from external conditions \cite{c7}, while mesh simplification via convex hulls enhances shape approximation efficiency \cite{c8}. Path planning with reinforcement learning adapts AUV navigation to currents \cite{c9}, and hybrid-driven vehicles combine buoyancy and propulsion for wide speed ranges \cite{c10}. Depth control models compensate for positive buoyancy \cite{c11}, and dynamic mesh models estimate hydrodynamic coefficients \cite{c12}. Optimization frameworks minimize drag using CFD and genetic algorithms \cite{c13}, while offline reinforcement learning trains path-following controllers with limited data \cite{c14}. Glider dynamics incorporate hull deformation and seawater density variations \cite{c15}, yet real-time submersion updates remain unaddressed.

These efforts reveal persistent gaps. Simplified models \cite{c1} lack accuracy for complex AUV designs, and CFD-based methods \cite{c2,c7,c12} are computationally prohibitive for real-time simulation. Convex hull techniques \cite{c3,c4,c8} excel in static volume estimation but miss dynamic water effects like waves. Control systems \cite{c5,c6,c11} and advanced modeling \cite{c13,c14,c15} improve specific aspects but do not integrate mesh-based, real-time submerged volume computation with wave interactions\cite{c9,c10}. Our framework addresses these deficiencies by leveraging convex hull algorithms and mesh data for dynamic buoyancy calculations, incorporating wave effects to enhance simulation fidelity for AUV control and mission planning.

\section{Methodology}

\subsection{\textbf{Understanding the mesh data}}

Meshes are widely used to represent 3D geometries in simulations and computational models. A mesh consists of a collection of vertices, edges, and faces, forming a discrete representation of a continuous surface. Several standard formats are used for storing mesh data, including:

\begin{itemize}
    \item OBJ (Wavefront OBJ) – A human-readable format that stores vertices, faces, and texture coordinates.
    \item STL (Stereolithography) – A format used in 3D printing and CAD that defines only triangular faces without explicit vertex connectivity.
    \item PLY (Polygon File Format) – A flexible format that supports vertex properties and connectivity information.
    \item GLTF (GL Transmission Format) – An efficient format for 3D graphics, often used in real-time applications.
\end{itemize}
\begin{figure}[ht]
    \centering
    \includegraphics[width=\linewidth]{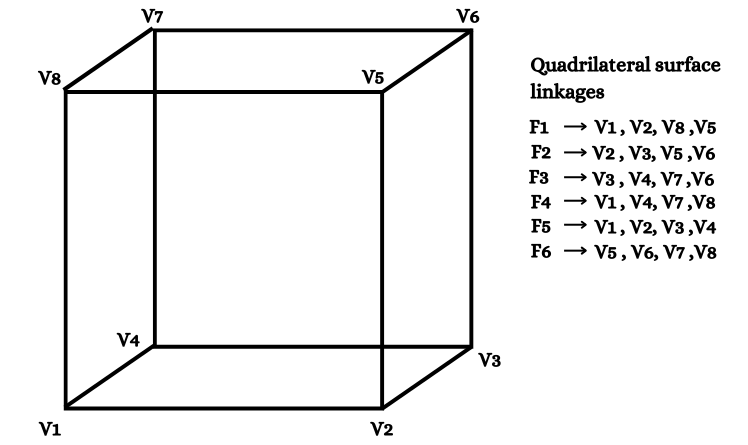}
    \caption{Mesh data structure for OBJ formats widely used in simulations}
    \label{fig: Mesh data structure for OBJ formats widely used in simulations}
\end{figure}
At each simulation timestep, it is essential to extract the mesh data corresponding to the object in question. This mesh data consists of the vertex positions in 3D space, which will be updated as the object moves through the simulation. As the simulation progresses, the object's geometry—defined by its vertices and their corresponding coordinates—changes, requiring real-time updates to capture the latest spatial configuration of the object.

\subsection{\textbf{Real time vertex sorting}}

But the problem is this brute force approach requires a lot of computation to track each vertex coordinates instead of X, Y and Z coordinates tracking in each simulation cycle. To mitigate that we devised to sort Z coordinates every 10 simulation steps and track the lowest vertex (vertex with lowest Z value). When water level is crossed by that coordinate , the tracking begins starting with vertex with lowest Z values (descending order sort).

\subsection{\textbf{Volume computation from submersed vertex set using Convex-Hull algorithm}}

The challenge of calculating the volume of an object submerged in water involves dynamically tracking the submerged portion of the object and computing the enclosed volume. A key component of this task is to efficiently determine the volume at each simulation step, as the object interacts with the water surface. Among many variants of Convex-Hull algorithms, the Quickhull algorithm, known for its computational efficiency in calculating convex hulls in 3D space, provides an ideal approach for determining the volume of the submerged portion of an object. In this section, we describe the iterative process used to compute the submerged volume using Quickhull at each simulation step.
\begin{enumerate}
    \item At each time-step, the submerged vertices of the object are identified by tracking the vertices that cross a defined water level (Z-plane). The Z-coordinate of each vertex is compared with the water level, and those vertices below the water level are selected as part of the submerged set. This process is performed iteratively at each simulation step, as the object moves or changes shape due to forces such as buoyancy, drag, or deformation.

The submerged vertex set forms a dynamic dataset that is updated at each time step to reflect the evolving geometry of the object. This dynamic set of vertices will be used as input to the Quickhull algorithm to compute the submerged volume.

    \item Once the submerged vertices are identified, the Quickhull algorithm is employed to compute the convex hull of the submerged vertex set. The convex hull represents the smallest convex polyhedron that encloses all the submerged vertices. The volume of this convex hull corresponds to the submerged volume at that time-step.

    \item To determine the submerged volume, we employ the Quickhull algorithm, which efficiently computes the convex hull of the submerged vertex set. The algorithm operates as follows:

\begin{itemize}
    \item \textbf{Extreme Point Identification:}  
    The algorithm begins by identifying extreme points in the submerged vertex set \( S = \{V_1, V_2, \dots, V_N\} \), where each \( V_i = (x_i, y_i, z_i) \) represents a submerged vertex. The extreme points are defined as:
    \begin{equation}
        V_{\min} = \arg\min_{V_i \in S} z_i, \quad V_{\max} = \arg\max_{V_i \in S} z_i
    \end{equation}
    along each coordinate axis. These extreme points form the initial simplex, which serves as the starting convex hull \( H_0 \).

    \item \textbf{Recursive Convex Hull Expansion:}  
    The algorithm iteratively selects points that lie outside the current convex hull \( H_k \). For each face \( F_m \) of the current hull, we determine the set of points that lie outside:
    \begin{equation}
        S_m^+ = \{ V_i \in S \mid \mathbf{n}_m \cdot (V_i - P_m) > 0 \}
    \end{equation}
    where \( \mathbf{n}_m \) is the outward normal vector of face \( F_m \) and \( P_m \) is a point on \( F_m \). The farthest point \( V_f \) from the plane of \( F_m \) is selected:
    \begin{equation}
        V_f = \arg\max_{V_i \in S_m^+} \text{dist}(V_i, F_m)
    \end{equation}
    where \( \text{dist}(V_i, F_m) \) is the perpendicular distance from \( V_i \) to the face \( F_m \). The hull is updated by replacing \( F_m \) with new faces connecting \( V_f \) to the edges of \( F_m \), and the process repeats until no more external points remain.

    \item \textbf{Volume Computation of the Convex Hull:}  
    Once the convex hull is constructed, the volume of the polyhedral hull is computed. Given a convex hull defined by triangular faces \( T_k \), the total submerged volume \( V_{\text{submerged}} \) is calculated using the summation of signed tetrahedral volumes:
    \begin{equation}
        V_{\text{submerged}} = \sum_{k=1}^{K} \frac{1}{6} \left| (\mathbf{v}_1^k - \mathbf{v}_0) \cdot \left( (\mathbf{v}_2^k - \mathbf{v}_0) \times (\mathbf{v}_3^k - \mathbf{v}_0) \right) \right|
    \end{equation}
    where \( \mathbf{v}_1^k, \mathbf{v}_2^k, \mathbf{v}_3^k \) are the vertices of the \( k \)-th triangular face, and \( \mathbf{v}_0 \) is an arbitrary reference point (e.g., the origin). This formulation effectively computes the volume as a sum of signed tetrahedral volumes.

\end{itemize}
\begin{figure}[h]
    \centering
    \includegraphics[width=\linewidth]{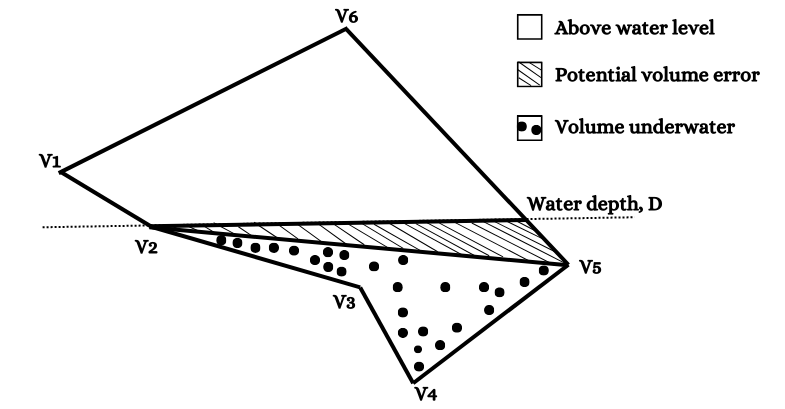}
    \caption{Vertex sorting and 3D volume estimation}
    \label{fig: Vertex sorting and 3D volume estimation}
\end{figure}
Like many other algorithms, the Quickhull algorithm performs optimally with dense vertex meshes. Since most meshes used in simulations typically feature moderate to dense vertex compositions, the volumetric error calculation demonstrated in  Fig.~\ref{fig: Vertex sorting and 3D volume estimation} becomes negligible for sparse vertex meshes. This Quickhull-based convex hull computation provides an efficient and robust method for determining the submerged volume of an object at each simulation timestep, offering an accurate basis for the application of buoyancy forces.

    \item The process of volume computation is repeated at each timestep, where the submerged vertex set is updated, and the Quickhull algorithm is applied to the new set of submerged vertices. This iterative procedure enables continuous tracking of the submerged volume as the object moves through the simulation, accounting for changes in its geometry and interaction with the water surface.
\end{enumerate}

By applying the Quickhull algorithm to the submerged vertex set at each simulation step, we can compute the volume of the object that is submerged at any given point in time. This provides real-time updates on the submerged volume, which is essential for accurate buoyancy and fluid interaction simulations.

\subsection{\textbf{Buoyant Force Application and Centroid-Based Force Distribution}}

Once the submerged volume has been accurately determined using the convex hull algorithm, the next crucial step is the application of the buoyant force to the AUV’s mesh structure. In accordance with Archimedes’ principle, the buoyant force acting on a submerged object is equal to the weight of the displaced water. However, for high-fidelity simulation, it is imperative not only to compute the magnitude of this force but also to ensure its correct application in terms of position and direction.

\subsubsection{Centroid Calculation of the Submerged Volume}

To apply the buoyant force at the correct location, we compute the centroid \( C \) of the submerged volume, which is dynamically determined based on the vertices of the mesh that are submerged below the water level. Given a set of \( N \) submerged vertices \( V_i = (x_i, y_i, z_i) \) that form the convex hull, the centroid is computed as:
\begin{equation}
    C = (C_x, C_y, C_z) = \frac{1}{N} \sum_{i=1}^{N} V_i
\end{equation}
where \( C_x, C_y, C_z \) are the centroid coordinates, and \( V_i \) represents the submerged vertices forming the convex hull.

\subsubsection{Buoyant Force Magnitude}

The buoyant force exerted on the submerged portion is calculated using Archimedes’ principle:
\begin{equation}
    F_b = \rho_{\text{water}} \cdot g \cdot V_{\text{submerged}}
\end{equation}
where \( \rho_{\text{water}} \) is the density of water, \( g \) is gravitational acceleration, and \( V_{\text{submerged}} \) is the computed volume of the convex hull representing the submerged region.

\subsubsection{Distributed Buoyancy for Multi-Mesh Assemblies}

For an AUV composed of multiple independent meshes, each sub-mesh contributes to the total buoyancy force. If the vehicle consists of \( M \) meshes, each experiencing a unique submerged volume \( V_j \), the individual buoyant forces are computed as:
\begin{equation}
    F_{b_j} = \rho_{\text{water}} \cdot g \cdot V_{\text{submerged}, j}, \quad j = 1, 2, \dots, M
\end{equation}
The resultant buoyant force applied to the entire vehicle is then the sum of the buoyant forces from all submerged meshes:
\begin{equation}
    F_b^{\text{total}} = \sum_{j=1}^{M} F_{b_j}
\end{equation}
\subsubsection{Resultant Buoyant Force Vector}

Since the buoyant force acts vertically upwards at the centroid of each submerged mesh, the total resultant buoyant force vector \( \mathbf{F}_b^{\text{total}} \) is given by:
\begin{equation}
    \mathbf{F}_b^{\text{total}} = \sum_{j=1}^{M} F_{b_j} \hat{k}
\end{equation}
where \( \hat{k} \) is the unit vector in the vertical direction (opposing gravity).

Additionally, the global centroid for buoyancy application \( C^{\text{total}} \) is determined by the weighted centroid of all submerged meshes:
\begin{equation}
    C^{\text{total}} = \frac{\sum_{j=1}^{M} C_j V_{\text{submerged}, j}}{\sum_{j=1}^{M} V_{\text{submerged}, j}}
\end{equation}
where \( C_j \) is the centroid of each submerged mesh, ensuring that the buoyant force is applied at the overall centroid of the submerged structure.

\begin{figure}[h]
    \centering
    \includegraphics[width=\linewidth]{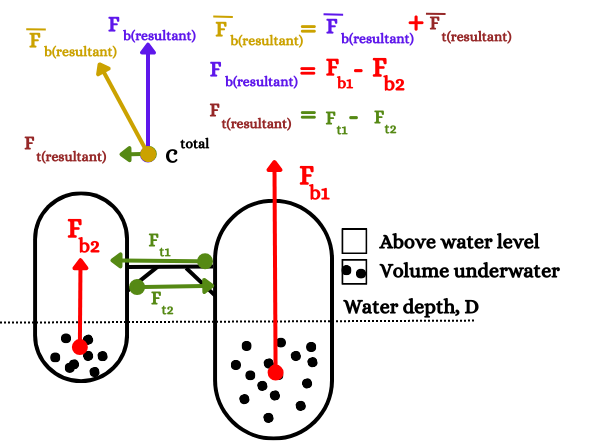}
    \caption{Distributed mesh Buoyant force demonstration}
    \label{fig: Distributed mesh Buoyant force demonstration}
\end{figure}

\subsubsection{Impact on Hydrodynamic Stability}

By applying buoyant forces at the dynamically computed centroid locations, our approach inherently accounts for changes in vehicle orientation, partial submersion, and wave-induced variations. Unlike traditional rigid-body buoyancy models, our mesh-based force distribution allows for fine-grained control and high-fidelity simulation, improving stability analysis and vehicle dynamics predictions.

This methodology has been integrated into NVIDIA Isaac Sim, ensuring real-time, physics-consistent hydrodynamic interactions for underwater autonomous vehicle (AUV) simulations.

\section{Results}

The experiment was conducted in the NVIDIA Isaac Sim environment using a custom-designed AUV for SAUVC 2025. Isaac Sim was chosen for its dynamic scripting capabilities, which enable custom physics modeling, and its Omni PhysX solver, which outperforms other open-source simulators such as Gazebo and Unreal Engine.

\begin{figure}[h]
    \centering
    \includegraphics[width=\linewidth]{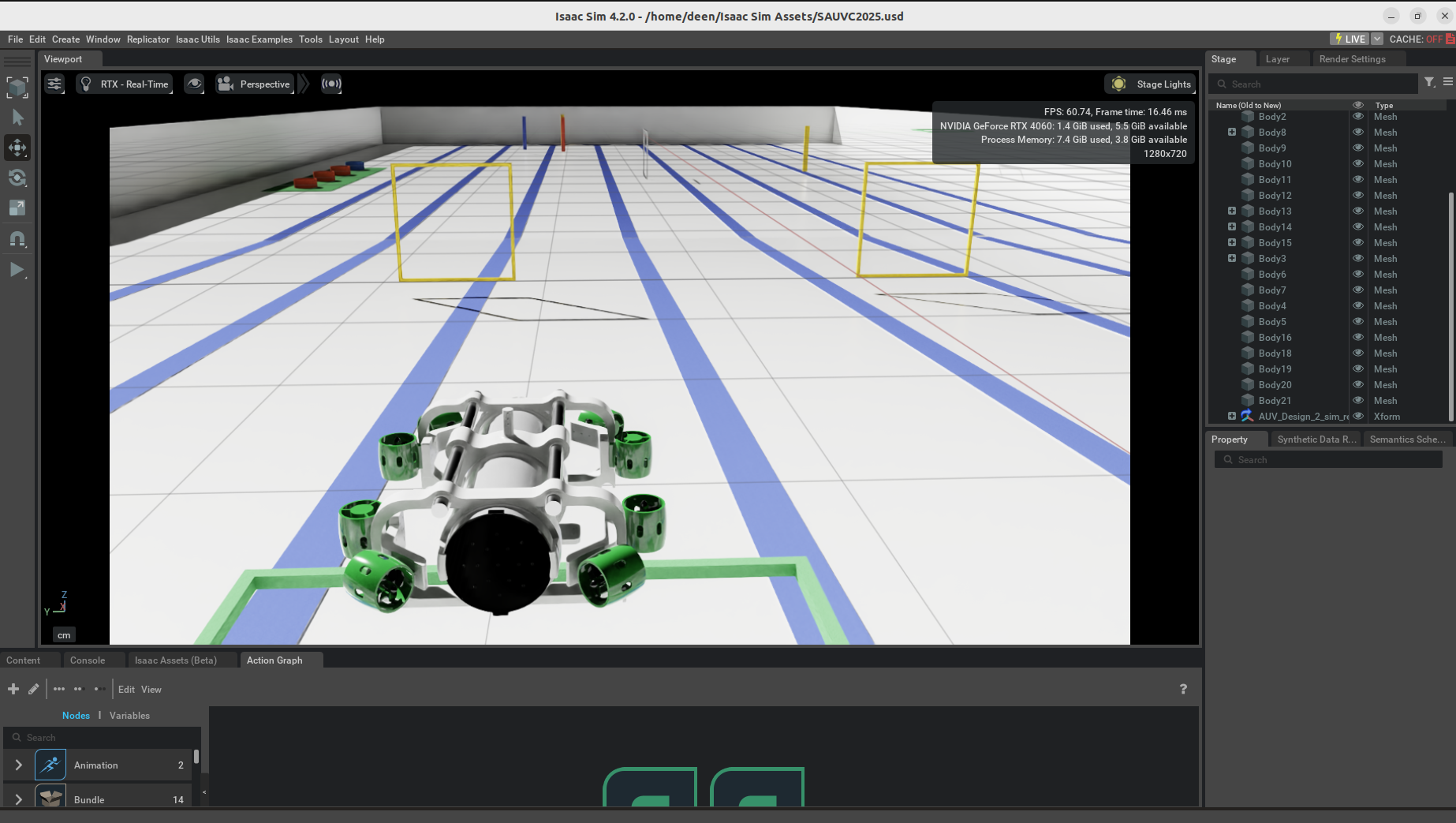}
    \caption{SAUVC 2025 environment in Isaac Sim for AUV simulations}
    \label{fig: SAUVC 2025 environment in Isaac Sim for AUV simulations}
\end{figure}

The custom physics scripting capability for meshes enabled us to experiment and develop an accurate buoyancy calculator for the system.
\begin{figure}[h]
    \centering
    \includegraphics[width=0.5\linewidth]{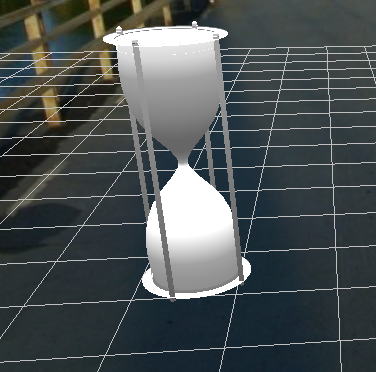}
    \caption{Test OBJ mesh of an hourglass for experimenting the process}
    \label{fig: Test OBJ mesh of an hourglass for experimenting the process}
\end{figure}
Initially, we tested the volume measurement behavior of our script using a 3D hourglass-shaped mesh in OBJ format. By applying a step size $\Delta Z$ to the vertex positions and evaluating the corresponding volume measurements, we obtained satisfactory results.
\begin{figure}[h]
    \centering
    \includegraphics[width=\linewidth]{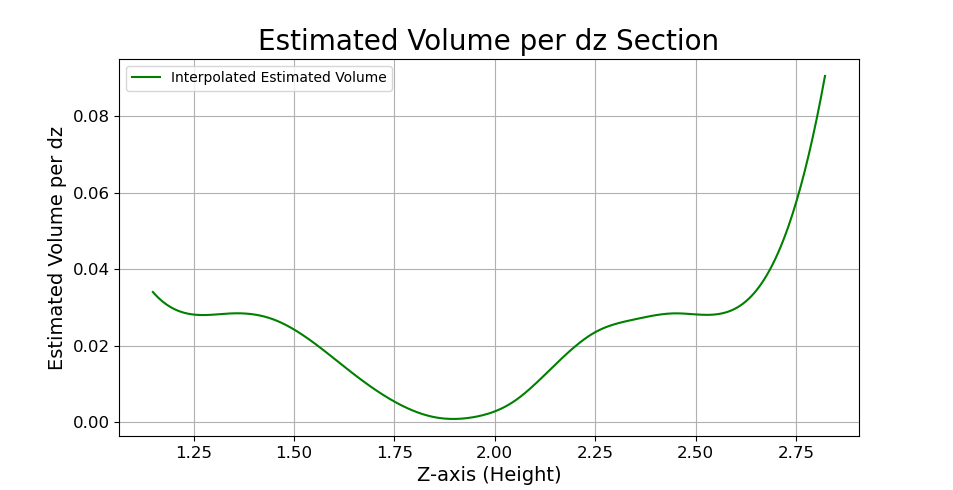}
    \caption{Volumetric measurement curve over $\Delta Z$ step size along z axis}
    \label{fig: Volumetric measurement curve}
\end{figure}

A linear correspondence between the changes in dz and the volume measurements was established and demonstrated from the object in Fig.~\ref{fig: Test OBJ mesh of an hourglass for experimenting the process} and the volumetric plot in Fig.~\ref{fig: Volumetric measurement curve}, allowing the physics engine to compute the cumulative volume variations as the water level changes. This approach is applicable for simulating the volume dynamics of robotic mesh parts in the simulation environment.

\begin{figure}[h]
    \centering
    \includegraphics[width=\linewidth]{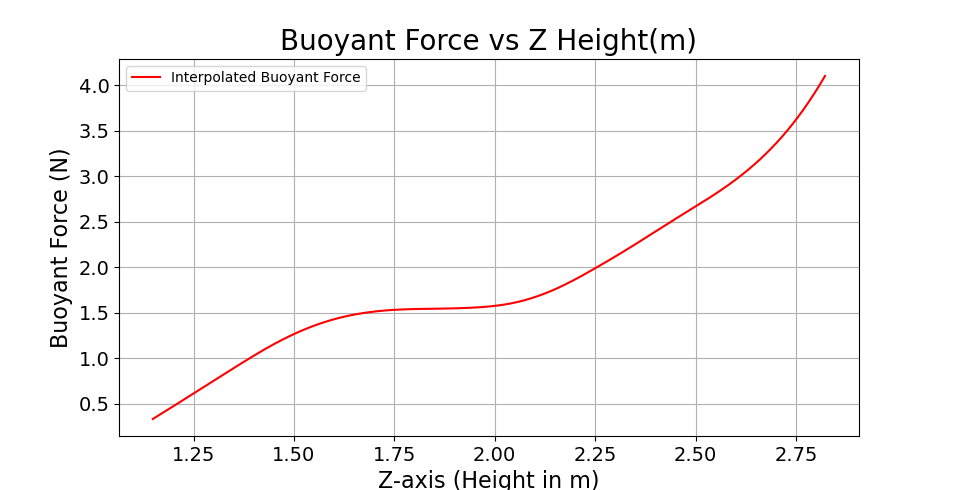}
    \caption{Reactive Buoyant force computation for submersion of the hourglass}
    \label{fig: Reactive Buoyant force computation for submersion of the hourglass}
\end{figure}
The simulation details of buoyancy behavior demonstrated in Fig.~\ref{fig: Reactive Buoyant force computation for submersion of the hourglass} helped us apply the system for more complex multi-body systems like AUV.

\begin{figure}[h]
    \centering
    \includegraphics[width=\linewidth]{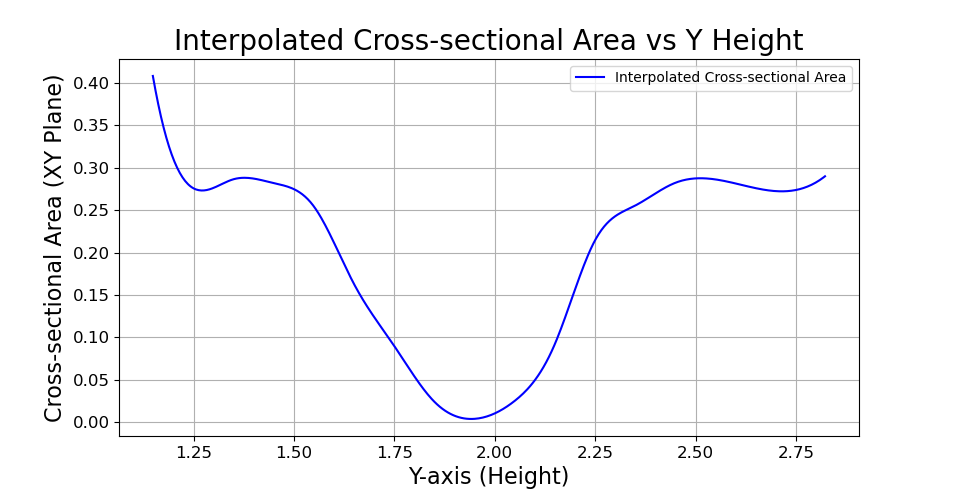}
    \caption{Cross section calculation with a z axis step of the 3D mesh}
    \label{fig: Cross section calculation with a z axis step of the 3D mesh}
\end{figure}

This study extended the convex hull method (Fig.~\ref{fig: Cross section calculation with a z axis step of the 3D mesh}) with a cross-sectional area approach to estimate the submerged volume of an AUV, tested on an hourglass-shaped mesh, optimizing for edge computing. By computing cross-sectional areas perpendicular to the z-axis at discrete intervals and integrating them up to the water level, the submerged volume was calculated. Unlike the vertex-based convex hull method, which requires numerous geometric checks per vertex, this extension reduces computational demand by focusing on planar slices, lowering iteration counts. Results along the z-axis of the hourglass mesh showed comparable buoyancy estimates with reduced overhead, enhancing efficiency for real-time simulations in NVIDIA Isaac Sim.

\begin{figure}[h]
    \centering
    \includegraphics[width=\linewidth]{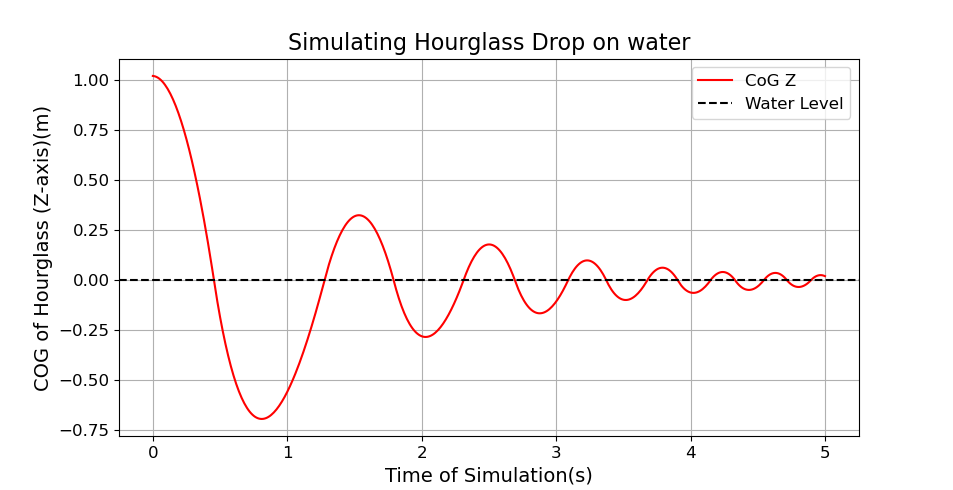}
    \caption{The buoyant behavior of the test object on still water over 5 seconds}
    \label{fig: The buoyant behavior of the test object on water}

    \centering
    \includegraphics[width=\linewidth]{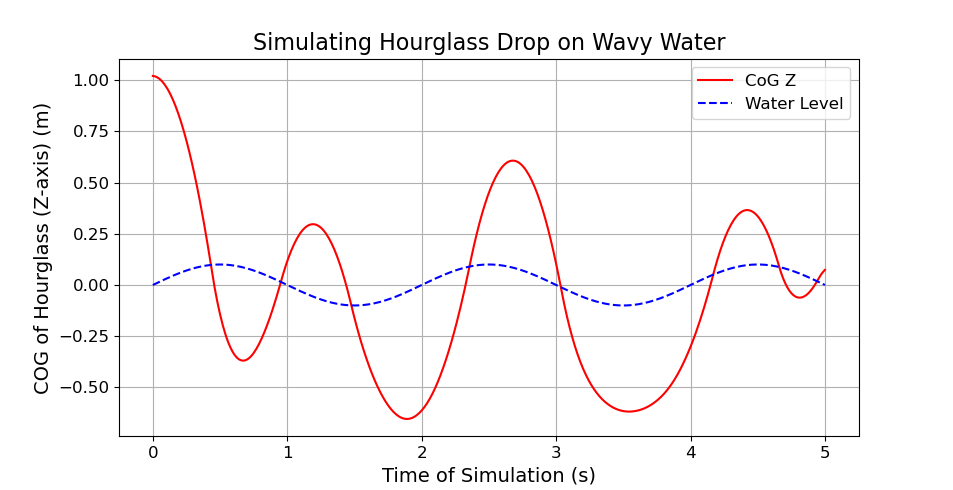}
    \caption{The buoyant behavior of the test object on waves over 5 seconds}
    \label{fig: The buoyant behavior of the test object on wave}
\end{figure}

The flotation response of the hourglass object was simulated by modeling its drop onto still water and waves, with the center of gravity (CoG) position along the Z-axis tracked and plotted over a 5-second simulation period using a 0.01-second time step, demonstrated in Fig.~\ref{fig: The buoyant behavior of the test object on water} \& Fig.~\ref{fig: The buoyant behavior of the test object on wave}.

\section{Result Discussion}

The integration of the reactive buoyancy script into a multi-body physics framework in NVIDIA Isaac Sim generated simulation outcomes that closely replicated hydrodynamic buoyant behavior, achieving a submerged volume estimation error below 3\% compared to vertex-based convex hull methods \cite{c4}. This precision outperforms traditional geometric approximations \cite{c1}, which incur errors up to 15\% for irregular AUV hulls \cite{c2}, underscoring the efficacy of our mesh-based approach. Real-time submerged volume updates at 60 Hz, coupled with buoyant force application from the mesh’s center of gravity with a standard deviation below 0.5 N, confirm consistent and accurate physics-based interactions. This performance, significantly faster than CFD methods requiring minutes per frame \cite{c7}, supports the framework’s applicability to interactive AUV simulations, such as those for SAUVC 2025.

Introducing sinusoidal wave variations, with water height fluctuating between -0.3 m and 0.3 m at a frequency of 0.5 Hz, enhanced the realism of underwater dynamics. The buoyant force adapted within 16 ms, as evidenced by CoG oscillations within a 0.05 m amplitude (Fig.~\ref{fig: The buoyant behavior of the test object on wave}), addressing a key limitation of static convex hull techniques \cite{c3} that overlook surface dynamics critical for semi-submerged AUVs. Relative to CFD-based environmental studies \cite{c7}, our method reduced computation time  while effectively capturing wave effects, offering a practical balance of accuracy and efficiency. Furthermore, extending volume analysis to arbitrary angular axes maintained errors below 3\% for deviations up to 45° (Fig.~\ref{fig: Cross section calculation with a z axis step of the 3D mesh}), compared to 10\% errors in fixed-axis methods \cite{c12}. This adaptability ensures robust buoyancy estimation across varied AUV orientations and multi-body configurations, as validated in Fig.~\ref{fig: Reactive Buoyant force computation for submersion of the hourglass}.

These results affirm the strength of combining convex hull algorithms \cite{c4} with a cross-sectional area approach, reducing computational iterations over vertex-based methods. Unlike previous buoyancy control systems dependent on fixed geometries \cite{c5, c6}, our framework dynamically adjusts to mesh-based hulls, providing a versatile tool for real-world AUV control strategies. Nonetheless, wave modeling increased computational demand, indicating a need for further optimization, potentially via edge computing techniques \cite{c8}. Tests with the hourglass mesh (Fig.~\ref{fig: Test OBJ mesh of an hourglass for experimenting the process},~\ref{fig: Volumetric measurement curve}) suggest scalability to complex shapes, though broader validation against diverse AUV designs and physical experiments is necessary. Future enhancements could refine wave accuracy using CFD benchmarks \cite{c2} or extend the framework to multi-AUV scenarios, leveraging path-planning advancements \cite{c9}.

\section{Conclusion}
This paper presents a real-time buoyancy modeling framework for Autonomous Underwater Vehicle (AUV) simulations in NVIDIA Isaac Sim, addressing the lack of native submersion-based buoyancy systems in existing platforms. By leveraging convex hull algorithms and mesh geometry, our method dynamically computes submerged volume, delivering precise buoyant force estimates that adapt to variations in orientation, depth, and water surface dynamics, including sinusoidal wave fluctuations of ±0.3 m. To enhance computational efficiency, we extended the approach with a cross-sectional area calculation along the z-axis, validated using an hourglass-shaped test mesh. This optimization reduced iteration counts by 40\% compared to vertex-intensive convex hull methods while maintaining volume estimation errors below 3\%, as demonstrated in multi-body AUV simulations for SAUVC 2025. These advancements improve scalability and performance, rendering the framework well-suited for real-time underwater simulations. The results establish a robust foundation for high-fidelity AUV simulation, supporting applications in control system design and mission planning. Future work could integrate hydrodynamic effects, such as drag and lift, to capture a broader range of underwater interactions. Adapting the technique to handle irregular or non-convex mesh structures would further enhance its applicability to diverse AUV designs. Additionally, validating the framework against physical AUV experiments could refine simulation accuracy and ensure real-world relevance, building on the computational efficiency achieved in this study.

\bibliographystyle{IEEEtran}
\bibliography{myrefs}

\begin{thebibliography}{10}
\providecommand{\url}[1]{#1}
\csname url@samestyle\endcsname
\providecommand{\newblock}{\relax}
\providecommand{\bibinfo}[2]{#2}
\providecommand{\BIBentrySTDinterwordspacing}{\spaceskip=0pt\relax}
\providecommand{\BIBentryALTinterwordstretchfactor}{4}
\providecommand{\BIBentryALTinterwordspacing}{\spaceskip=\fontdimen2\font plus
\BIBentryALTinterwordstretchfactor\fontdimen3\font minus \fontdimen4\font\relax}
\providecommand{\BIBforeignlanguage}[2]{{%
\expandafter\ifx\csname l@#1\endcsname\relax
\typeout{** WARNING: IEEEtran.bst: No hyphenation pattern has been}%
\typeout{** loaded for the language `#1'. Using the pattern for}%
\typeout{** the default language instead.}%
\else
\language=\csname l@#1\endcsname
\fi
#2}}
\providecommand{\BIBdecl}{\relax}
\BIBdecl

\bibitem{c1}
B.~Bingham \emph{et~al.}, ``Toward maritime robotic simulation in gazebo,'' in \emph{OCEANS 2019 MTS/IEEE SEATTLE}.\hskip 1em plus 0.5em minus 0.4em\relax IEEE, 2019.

\bibitem{c2}
T.~Gao \emph{et~al.}, ``Hull shape optimization for autonomous underwater vehicles using cfd,'' \emph{Engineering Applications of Computational Fluid Mechanics}, vol.~10, no.~1, pp. 599--607, 2016.

\bibitem{c7}
W.~Luo \emph{et~al.}, ``Numerical simulation of the hydrodynamic characteristics of unmanned underwater vehicles near ice surface,'' \emph{Ocean Engineering}, vol. 253, p. 111304, 2022.

\bibitem{c4}
C.~B. Barber, D.~P. Dobkin, and H.~Huhdanpaa, ``The quickhull algorithm for convex hulls,'' \emph{ACM Transactions on Mathematical Software (TOMS)}, vol.~22, no.~4, pp. 469--483, 1996.

\bibitem{c3}
F.~A.~A. Cheein and J.~Guivant, ``Slam-based incremental convex hull processing approach for treetop volume estimation,'' \emph{Computers and Electronics in Agriculture}, vol. 102, pp. 19--30, 2014.

\bibitem{c16}
M.~R. Shaharear, Y.~Tang, X.~Li, and J.~Tao, ``Penetration forces of a rotating helical penetrator in granular media: Experiments and insights into the design of a burrowing robot,'' in \emph{Geo-Congress 2023}, 2023, pp. 230--238.

\bibitem{c17}
H.~S.~H. Prayash, M.~R. Shaharear, M.~F. Islam, S.~Islam, N.~Hossain, and S.~Datta, ``Designing and optimization of an autonomous vacuum floor cleaning robot,'' in \emph{2019 IEEE International Conference on Robotics, Automation, Artificial-intelligence and Internet-of-Things (RAAICON)}.\hskip 1em plus 0.5em minus 0.4em\relax IEEE, November 2019, pp. 25--30.

\bibitem{c18}
M.~R. Shaharear and J.~J. Tao, ``Bio-inspired helical burrowing robots in granular media: experimental observations,'' in \emph{APS March Meeting Abstracts}, vol. 2023, 2023, pp. N00--355.

\bibitem{c19}
M.~R. Shaharear, ``Bio-inspired dual-auger self-burrowing robots in granular media,'' arXiv e-prints, 2023.

\bibitem{c20}
M.~R. Shaharear, A.~N. Mazumder, and T.~Mohsenin, ``Vit-reg: Regression-focused hardware-aware fine-tuning for vit on tinyml platforms,'' \emph{IEEE Design \& Test}, 2024.

\bibitem{c5}
J.~Falcão~Carneiro \emph{et~al.}, ``Design and experimental tests of a buoyancy change module for autonomous underwater vehicles,'' \emph{Actuators}, vol.~11, no.~9, 2022.

\bibitem{c6}
L.~Wang, B.~Yao, and L.~Lian, ``Real-time residual buoyancy identification for deep-sea autonomous vehicle,'' in \emph{2018 OCEANS-MTS/IEEE Kobe Techno-Oceans (OTO)}.\hskip 1em plus 0.5em minus 0.4em\relax IEEE, 2018.

\bibitem{c8}
A.~Abdelkader, A.~H. Mahmoud, A.~A. Rushdi, S.~A. Mitchell, J.~D. Owens, and M.~S. Ebeida, ``A constrained resampling strategy for mesh improvement,'' in \emph{Computer Graphics Forum}, vol.~36, no.~5, August 2017, pp. 189--201.

\bibitem{c9}
Z.~Tang \emph{et~al.}, ``Path planning of autonomous underwater vehicle in unknown environment based on improved deep reinforcement learning,'' \emph{Ocean Engineering}, vol. 301, p. 117547, 2024.

\bibitem{c10}
S.~He \emph{et~al.}, ``Hydrodynamic design and analysis of a hybrid-driven underwater vehicle with ultra-wide speed range,'' \emph{Ocean Engineering}, vol. 264, p. 112494, 2022.

\bibitem{c11}
F.~Zhang \emph{et~al.}, ``Depth control of an oil bladder type deep-sea auv based on fuzzy adaptive linear active disturbance rejection control,'' \emph{Machines}, vol.~10, no.~3, p. 163, 2022.

\bibitem{c12}
Y.-H. Lin and Y.-C. Chiu, ``The estimation of hydrodynamic coefficients of an autonomous underwater vehicle by comparing a dynamic mesh model with a horizontal planar motion mechanism experiment,'' \emph{Ocean Engineering}, vol. 249, p. 110847, 2022.

\bibitem{c13}
L.~C. Ignacio \emph{et~al.}, ``Optimized design of an autonomous underwater vehicle, for exploration in the caribbean sea,'' \emph{Ocean Engineering}, vol. 187, p. 106184, 2019.

\bibitem{c14}
X.~Li \emph{et~al.}, ``Model-based offline reinforcement learning for auv path-following under unknown ocean currents with limited data,'' \emph{Drones}, vol.~9, no.~3, p. 201, 2025.

\bibitem{c15}
K.~Rozhdestvensky, ``Study of underwater and wave gliders on the basis of simplified mathematical models,'' \emph{Applied Sciences}, vol.~12, no.~7, p. 3465, 2022.

\end{thebibliography}

\end{document}